\documentclass[letterpaper, 10 pt, conference]{ieeeconf}            

\IEEEoverridecommandlockouts
                                                          
\usepackage{amsmath,amsfonts}
\usepackage{array}
\usepackage{textcomp}
\usepackage{stfloats}
\usepackage{url}
\usepackage{verbatim}
\usepackage{graphicx}
\usepackage{cite}
\usepackage[utf8]{inputenc}
\usepackage{hyperref}

\usepackage{subcaption}
\usepackage{bm}
\usepackage{rotating}
\usepackage{multirow}
\usepackage{hhline}

\usepackage{soul}
\usepackage{afterpage}
\usepackage{color}
\usepackage{latexsym}
\usepackage{amssymb}
\usepackage{mathtools}
\usepackage{pdflscape}
\usepackage{enumerate}
\let\labelindent\relax
\usepackage[inline]{enumitem}
\usepackage{xspace} 
\usepackage{float}
\usepackage{booktabs}
\usepackage{diagbox}

\usepackage{tabu}
\usepackage{tabularx}
\usepackage{colortbl}

\newcommand\rowtag[2]{#1\def\@currentlabel{#1}\label{#2}}

\usepackage{pifont} 
\usepackage{bbding} 

\usepackage[dvipsnames,table]{xcolor}
\usepackage[normalem]{ulem}
\usepackage{makecell}
\usepackage{algorithm2e}
\usepackage{multicol}
\usepackage[capitalise]{cleveref}
\usepackage[percent]{overpic}
\usepackage{tikz}
\usetikzlibrary{arrows.meta,arrows,tikzmark,calc,backgrounds,decorations.pathreplacing,calligraphy}

\usepackage{cuted}

\usepackage[per-mode=fraction]{siunitx}
\usepackage{comment}
\usepackage[normalem]{ulem}
\usepackage{xspace}
\usepackage{lastpage}
\usepackage{fancyhdr}

\DeclareSIUnit\px{px}
\DeclareSIUnit\fps{fps}

\definecolor{OliveGreen}{RGB}{0,200,25}
\newcommand{\red}[1]{\textcolor{red}{#1}}

\newcommand{\darkgreen}[1]{\textcolor{OliveGreen}{#1}}

\newcommand{\ie}{i.\,e.,\xspace}
\newcommand{\eg}{e.\,g.,\xspace}

\newcommand{\armarVI}{\mbox{ARMAR-6}\xspace}

\newcommand{\armarDE}{\mbox{ARMAR-DE}\xspace}

\newcommand{\ackJuBoteuROBIN}{This work was supported by the Carl Zeiss Foundation under the project JuBot and the European Union's Horizon Europe Framework Programme under grant agreement No 101070596 (euROBIN).}

\newif\iffinal

\newcommand{\replaced}[2]{%
	\iffinal%
	#2%
	\else%
	\red{\ifmmode\text{\sout{\ensuremath{#1}}}\else\sout{#1}\fi}\darkgreen{#2}%
	\fi%
}
\newcommand{\removed}[1]{%
	\iffinal%
	\else%
	\red{\ifmmode\text{\sout{\ensuremath{#1}}}\else\sout{#1}\fi}%
	\fi%
}

\definecolor{cadmiumyellow}{rgb}{1.0, 0.96, 0.0}
\definecolor{canaryyellow}{rgb}{1.0, 0.94, 0.0}
\definecolor{bananayellow}{rgb}{1.0, 0.88, 0.21}
\definecolor{yellow(process)}{rgb}{1.0, 0.94, 0.0}
\definecolor{applegreen}{rgb}{0.55, 0.71, 0.0}
\definecolor{capri}{rgb}{0.0, 0.75, 1.0}
\definecolor{corn}{rgb}{0.98, 0.93, 0.36}
\definecolor{cornflowerblue}{rgb}{0.39, 0.58, 0.93}
\definecolor{darkblue}{rgb}{0.0, 0.0, 0.55}

\definecolor{c_ndf}{HTML}{f8cb7f}
\definecolor{c_nift}{HTML}{f89588}
\definecolor{c_mimo3}{HTML}{63b2ee}
\definecolor{c_mimo4s}{HTML}{7cd6cf}

\definecolor{mesh_blue}{HTML}{2399d7}
\definecolor{mesh_green}{HTML}{42eb72}
\definecolor{mesh_hand}{HTML}{cf8c62}

\usepackage{tikz-imagelabels}

\DeclareCaptionLabelFormat{custom}
{%
}
\DeclareCaptionLabelSeparator{custom}{--}
\DeclareCaptionFormat{custom}
{%
}

\usepackage{mathrsfs}
\newcommand{\degree}{^{\circ}}

\newcommand{\set}[1]{\{#1\}}

\DeclareMathOperator*{\argmax}{arg\,max}
\DeclareMathOperator*{\argmin}{arg\,min}

\DeclareRobustCommand\sampleline[1]{% line style
    \tikz\draw[#1] (0,0) (0,\the\dimexpr\fontdimen22\textfont2\relax) -- (1em,\the\dimexpr\fontdimen22\textfont2\relax);%
}

\DeclareRobustCommand\inlinearrow[1]{ % arg: color
    \hspace{-0.7em}
    \tikz{
        \draw[-{to[length=1.5mm]},
        #1,line width=1.1pt](0,0) (0,\the\dimexpr\fontdimen22\textfont2\relax) -- (1em,\the\dimexpr\fontdimen22\textfont2\relax);
    }
    \hspace{-0.7em}
}

\DeclareRobustCommand\inlinepoint[1]{
    $\textcolor{#1}{\bullet}$
}

\usepackage{amsmath}
\usepackage{multirow}

\newenvironment{talign*}
 {\let\displaystyle\textstyle\csname align*\endcsname}
 {\endalign}

\newcolumntype{P}[1]{>{\centering\arraybackslash}p{#1}}
\newcolumntype{M}[1]{>{\centering\arraybackslash}m{#1}}

\newlength\imgwidth
\newlength\imgheight

\newsavebox{\subfigbox}

\newcommand{\zz}[1]{\textbf{#1}}

\newcommand{\mimolong}{\emph{Multi-feature Implicit Model}\xspace}
\newcommand{\mimo}{MIMO\xspace}
\newcommand{\mimoIII}{MIMO3\xspace}
\newcommand{\mimoIV}{MIMO4\xspace}
\newcommand{\ndf}{NDF\xspace}
\newcommand{\rndf}{R-NDF\xspace}
\newcommand{\nift}{NIFT\xspace}

\newcommand{\occ}{$\mathbf{\Phi}_{\text{occ}}$\xspace}
\newcommand{\sdf}{$\mathbf{\Phi}_{\text{sdf}}$\xspace}
\newcommand{\escf}{$\mathbf{\Phi}_{\text{escf}}$\xspace}
\newcommand{\cdd}{$\mathbf{\Phi}_{\text{cdd}}$\xspace}

\newcommand{\eqs}[1]{\scalebox{0.9}{#1}}
\newcommand{\objs}{\eqs{$\mathcal{O}_s$}\xspace}
\newcommand{\objt}{\eqs{$\mathcal{O}_t$}\xspace}
\newcommand{\objA}{\eqs{$\mathcal{O}_A$}\xspace}
\newcommand{\objAnew}{\eqs{$\bar{\mathcal{O}}_A$}\xspace}
\newcommand{\objB}{\eqs{$\mathcal{O}_B$}\xspace}
\newcommand{\objBnew}{\eqs{$\bar{\mathcal{O}}_B$}\xspace}
\newcommand{\dposeAB}{{}^A\mathbf{Z}_B}
\newcommand{\dposeABref}{{}^A\hat{\mathbf{Z}}_B}
\newcommand{\poseAB}{\eqs{$\dposeAB$}\xspace}
\newcommand{\poseABref}{\eqs{$\dposeABref$}\xspace}

\newcommand{\pclA}{\mathbf{P}^A}
\newcommand{\pclAr}{\mathbf{P}^A_r}

\newcommand{\pclAnew}{\bar{\mathbf{P}}^A_r}

\newcommand{\upPose}{$\mathsf{U}$\xspace}
\newcommand{\arbiPose}{$\mathsf{A}$\xspace}

\begin{document}

\title{\LARGE \bf Visual Imitation Learning of Task-Oriented \\ Object Grasping and Rearrangement}

\author{
    Yichen Cai$^{*}$,
    Jianfeng~Gao$^{*}$,
    Christoph Pohl,
    and~Tamim~Asfour
        \thanks{$^{*}$The authors contributed equally to the paper.}
    \thanks{\ackJuBoteuROBIN}% <-this % stops a space
    \thanks{The authors are with the Institute for Anthropomatics and Robotics, Karlsruhe Institute of Technology, Karlsruhe, Germany. E-mails: \mbox{ujfao@student.kit.edu}, \{jianfeng.gao, asfour\}@kit.edu}
    \thanks{This work has been submitted to the IEEE for possible publication. Copyright may be transferred without notice, after which this version may no longer be accessible.}
}

\makeatletter
\let\@oldmaketitle\@maketitle% Store \@maketitle
\renewcommand{\@maketitle}{\@oldmaketitle% Update \@maketitle to insert...
	% \vspace{-10.5ex}
}
\makeatother
\maketitle

\begin{abstract}
    Task-oriented object grasping and rearrangement are critical skills for robots to accomplish different real-world manipulation tasks. 
    However, they remain challenging due to partial observations of the objects and shape variations in categorical objects.
    In this paper, we propose the \mimolong (\mimo), a novel object representation that encodes \emph{multiple spatial features} between a point and an object in an implicit neural field. 
    Training such a model on multiple features ensures that it embeds the object shapes consistently in different aspects, thus improving its performance in object shape reconstruction from partial observation, shape similarity measure, and modeling spatial relations between objects. 
    Based on \mimo, we propose a framework to learn task-oriented object grasping and rearrangement from single or multiple human demonstration videos. 
    The evaluations in simulation show that our approach outperforms the state-of-the-art methods for multi- and single-view observations. Real-world experiments demonstrate the efficacy of our approach in one- and few-shot imitation learning of manipulation tasks.
\end{abstract}

%===============================================================================

% keywords: Deep Learning in Grasping and Manipulation, Imitation Learning, Perception for Grasping and Manipulation

\section{Introduction}
\label{sec:introduction}
\begin{figure}
    \centering
    
    \begin{subfigure}{0.48\textwidth}
            \begin{tikzpicture}
                \node (sim) at (-2.1, 0) {\includegraphics[width=0.48\textwidth]{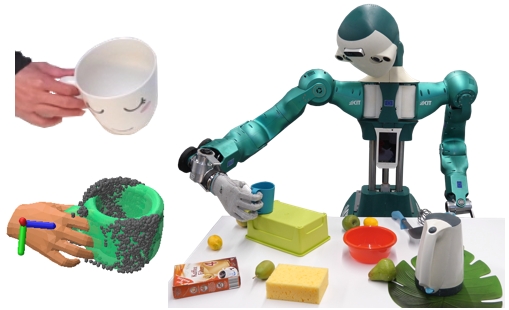}};
                \node (sim) at ( 2.1, 0) {\includegraphics[width=0.48\textwidth]{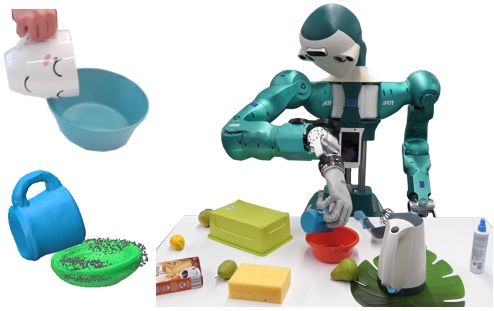} };
                \node at (-3.8, -0.2) {\scalebox{0.8}{$\mathbf{T}_g^d$}};
            \end{tikzpicture}
        \caption{Side Grasp and Pouring.}
        \label{subfig:side_g}
        \end{subfigure}
        
        \vspace{0.8em}
        
        \begin{subfigure}{0.48\textwidth}
            \centering
            \begin{tikzpicture}
                \node (sim) at (-2.1, 0) {\includegraphics[width=0.48\textwidth]{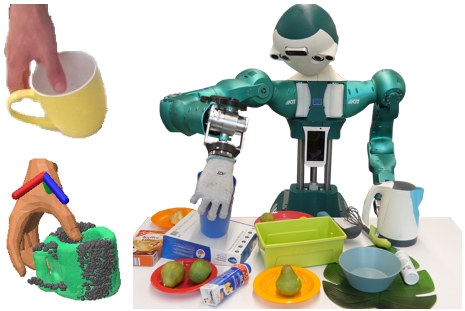}};
                \node (sim) at ( 2.1, 0) {\includegraphics[width=0.48\textwidth]{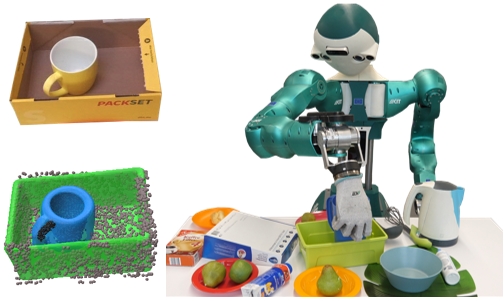}};
                \node at (-3.4, -0.2) {\scalebox{0.8}{$\mathbf{T}_g^d$}};
            \end{tikzpicture}
        \caption{Top-down Grasp and Placement.}
        \label{subfig:top_g}
    \end{subfigure}
    \vspace{-1em}
    \caption{Learning task-oriented object grasping and rearrangement from human demonstration videos of manipulation tasks. We illustrate two tasks: \subref{subfig:side_g} side picking a mug and pouring into a bowl; and \subref{subfig:top_g} top-down picking a mug and placing it into a container. For each task, we show the RGB image, the observed point clouds (\inlinepoint{black}), reconstructed object meshes (\raisebox{-1.0mm}{\sampleline{mesh_blue,line width=2mm}}\raisebox{-1.0mm}{\sampleline{mesh_green,line width=2mm}}), extracted hand mesh (\raisebox{-1.0mm}{\sampleline{mesh_hand,line width=2mm}}), grasp poses (\eqs{$\mathbf{T}_g^d$}), and the execution on a humanoid robot.}
    \vspace{-1.4em}
    \label{fig:teasor}
\end{figure}

Performing accurate manipulation tasks with everyday objects is an intricate problem that poses several challenges for robots. 
The robot must first find the optimal grasps for specific tasks and generate a suitable motion trajectory to achieve this configuration.
For instance, a side grasp by the mug handle is suitable for pouring water out of a mug (see \cref{subfig:side_g}), while a top grasp by the rim is more suitable when placing the mug into a container to avoid collision between the hand and the container (see \cref{subfig:top_g}). 
Additionally, suitable pose configurations of the mug relative to the bowl and the container are needed in such an object rearrangement task.

To generate task-oriented grasps, previous works~\cite{kokic2017affordance, detry2017task, nguyen2017object} have focused on training neural networks on large manually annotated datasets. Despite their performance, these approaches fail to generalize to novel objects with large shape variations. Moreover, manual annotation is costly and difficult to acquire.
In contrast, visual imitation learning (VIL) approaches like \cite{jin_generalizable_2022, gao2023k} provide efficient means to teach robots manipulation skills from human demonstrations and enable generalization to new scenarios with categorical objects.

This paper focuses on the line of works that utilize neural fields, \eg \cite {xie2022neural, simeonov2022neural, huang2023nift, simeonov2023se}, which implicitly encode object spatial properties.
Neural fields can be trained in a self-supervised manner by exploiting an inherent bias towards object classes, thus eliminating the need for manual annotation. 
This bias plays an important role in establishing dense 3D correspondences across categorical objects, enabling the adaptation of object manipulation skills to previously unseen object instances. 
However, these approaches require multiple views of the object, which are often unavailable in real-world applications. 
When presented with a partial view or categorical objects with large shape variations, these approaches may yield less precise grasp or object target poses, potentially resulting in collisions or unstable placement.

To address the above-mentioned challenges, we introduce the \mimolong (\mimo), which is designed to predict multiple spatial properties of a 3D point relative to an object. 

This enables our model to generate a richer descriptor space and thus more precise dense correspondences, which facilitates the accurate transfer of grasps and object target poses to new situations.
\mimo can also reconstruct object shapes when only a partial observation is available, which is beneficial for coping with task constraints defined on the hidden part of the object.
Leveraging \mimo's capabilities, we propose a framework that efficiently learns and generates task-oriented grasps from single or multiple human demonstration videos. 
Moreover, we use an evaluation network to predict the success probability of the generated grasps and refine them if necessary.

The contributions of this paper are twofold: 
\begin{enumerate*}[label=(\arabic*)] 
\item We propose the novel \mimolong (\mimo) that predicts multiple spatial features of a point relative to an object, which yields an informative point and pose descriptor space. It outperforms the state-of-the-art neural field methods in terms of dense correspondence, shape reconstruction, and pose transfer. The model can be trained in a self-supervised manner without relying on human annotations.
\item We integrate \mimo into visual imitation learning and propose a framework to efficiently learn, generate, and refine task-oriented grasps. We achieve one- and few-shot imitation learning and demonstrate a direct transfer of the learned manipulation tasks to categorical objects.
\end{enumerate*}

%===============================================================================

\section{Related Work}
\label{sec:relatedwork}
Deep learning-based methods for grasping have made significant progress in robotics thanks to advances in implicit object representation. Explicitly modeling the relevance of manipulation skills for a given task is important for generalization to novel situations. In this regard, we focus on implicit representation through neural fields, along with recent advancements in task-relevant grasping and manipulation.

\subsection{Neural Fields and Neural Descriptors}
\label{sec:sota_neural_fields}
Neural-fields-based approaches involve training neural networks to learn a continuous representation by predicting the physical and spatial properties of a 3D point relative to its surroundings~\cite{xie2022neural}. 
The learned representations, known as descriptors, are used in various tasks such as 3D reconstruction \cite{mescheder2019occupancy, park2019deepsdf} and manipulation \cite{pfrommer2021contactnets, karunratanakul2020grasping}. 
Leveraging dense correspondences in the descriptor space allows the transfer of manipulation skills between similar objects.
Previous works \cite{florence2018dense, chai2019multi, yang2021learning} used Convolutional Neural Networks to obtain pixel-wise descriptors for detecting correspondences from RGB images.
However, these approaches rely on visible 2D descriptors, which fail to account for task constraints on the hidden parts of the objects.
To overcome this limitation, \emph{Neural Descriptor Fields} (NDFs) \cite{simeonov2022neural} directly encode SE(3)-equivariant point and pose descriptors from the 3D point cloud of objects.
Although a richer descriptor space was proposed in \cite{huang2023nift} by leveraging the space coverage feature (SCF) \cite{zhao2016relationship}, it sacrificed the ability to reconstruct object shapes.
Despite their performance for grasp transfer in multi-view cases with a few demonstrations (5-10), the accuracy degenerates where only a partial view or a single demonstration is available.
In contrast, we train the implicit model to predict multiple spatial features of a point relative to an object, resulting in a more informative descriptor space while preserving the shape reconstruction capability.
We outperform the approaches presented in \cite{simeonov2022neural} and \cite{huang2023nift} in tasks such as shape similarities measure and pose transfer, especially with partial view. 
This also leads to better performance in one-shot imitation learning of manipulation tasks.

\subsection{Modeling Task Relevance}
\label{sec:sota_task_grasp}
In the context of task-oriented grasping, it is crucial to consider the modeling of task relevance as this enables the determination of grasp poses that are most conducive to the downstream task.
In previous works, semantic segmentation models have been trained to detect grasp affordance regions from either RGB images~\cite{kokic2017affordance, detry2017task, nguyen2017object,xu2021affordance} or 3D point clouds~\cite{ardon2019learning, monica2020point, jiang2021synergies, chen2022learning}.
However, these methods often rely on large annotated datasets, necessitating time-consuming manual annotation. Furthermore, they are tailored to grasping rather than object rearrangement tasks.
The former challenge is alleviated in \cite{fang2020learning, wen2022catgrasp} via self-supervision in simulation.
To address the latter, recent works focus on modeling task relevance using general 2D or 3D neural descriptors,
\eg 2D affordance regions~\cite{holomjova2023one,hadjivelichkov2023one} and 3D affordance maps~\cite{rashid2023language,kerr2023lerf}.
These neural descriptors measure shape similarity, enabling the modeling of task relevance and facilitating the transfer of task-relevant grasps, object poses, or regions to new scenarios.
However, the approaches in ~\cite{holomjova2023one} and \cite{hadjivelichkov2023one} are limited to top-down planar grasps, while multiple calibrated RGB images are required in~\cite{rashid2023language} and \cite{kerr2023lerf} for scene reconstruction, which is time-consuming and not always feasible.
In contrast, we use a novel neural pose descriptor derived from partial observations, which can be used for modeling task-oriented grasp distributions and downstream rearrangement tasks. 

\subsection{Category-Level Manipulation}
\label{sec:sota_cat_man} 
Previous works, like \cite{manuelli2019kpam, gao2021kpam, gao2021kpam20}, utilized semantic keypoints for transferring manipulation skills between categorical objects. 
However, overlooking the category-level inductive bias, these approaches necessitate extensive manual annotation for keypoint detection and careful assignment of keypoints for each task and object.
To address this problem, category-level non-rigid registration~\cite{rodriguez2018transferring, rodriguez2018transferring2,biza2023one} was proposed to reconstruct object shapes and infer object 6D poses.
However, these models face difficulties in transferring to objects with large shape variations.
Another line of work \cite{simeonov2022neural, huang2023nift} leverages category-level neural descriptors for transferring skills. 
However, these models assume one interacting object is known and fixed.
Relational-NDF (\rndf) \cite{simeonov2023se} relaxed this limitation by manually selecting keypoints and associated local frames in task-relevant regions.  
However, it faces challenges in predicting precise dense correspondences with partial observation, a limitation addressed by\cite{hidalgo2023anthropomorphic} through subtasks, including pose estimation, shape reconstruction, similarity measure, and grasp transfer.
Yet, each subtask demands a separate model. 
In contrast, we address partial observation by leveraging \mimo's capability in shape reconstruction.
This enhances the precision of task relevance and knowledge transfer for object grasping and rearrangement and allows the usage of \mimo for all tasks, offering an efficient solution for manipulation tasks. 
%===============================================================================

%===============================================================================
\begin{figure*}[t]
    \centering
    \tikzset{
        % Specifications for style
        base/.style = {rectangle, draw=black, minimum width=2mm, minimum height=6mm, fill=yellow!40},
        encoder/.style = {base, thick, minimum width=4mm, minimum height=2mm, rounded corners, fill=gray!20},
         block/.style = {base, fill=blue!30},
        block1/.style = {base, fill=red!20},
        block2/.style = {base, fill=yellow!20},
        block3/.style = {base, fill=green!20},
        s_encoder/.style = {},
    }
    % We need layers to draw the block diagram
    \tikzstyle{point}=[coordinate, on grid, ]
    \pgfdeclarelayer{background}
    \pgfdeclarelayer{foreground}
    \pgfsetlayers{background, main, foreground}
    \setlength\tabcolsep{0.5pt}
    \tabulinesep=0pt
    \begin{tabu}{M{0.40\textwidth}M{0.14\textwidth}M{0.25\textwidth}M{0.20\textwidth}}
        \multirow{2}{*}{
            \hspace{-15pt}
            \begin{subfigure}{\linewidth}
                \centering
                \begin{tikzpicture}
                    \node (pcl) {\includegraphics[width=0.1\textwidth]
                    {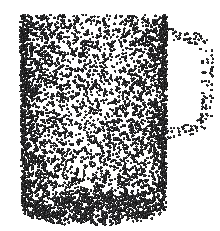}};
                    \node [below = 3mm of pcl, align=center, font=\small] at (pcl.center) 
                    {$\mathbf{P}$};
                    \node (point) [below = 6mm of pcl, align=center, font=\small] 
                    {$\mathbf{x \in \mathbb{R}^{3}}$};
                    
                    \node (encoder) [encoder, right = 4mm of pcl, align=center, font=\small]
                    {$\mathbf{\epsilon(P)}$};
                    
                    \node (concat1) [below right = 1mm and 0.7mm of encoder, inner sep=0] 
                    {$\mathbf{\oplus}$};
            
                    % % decoder
                    \node (s1) [block, right = 1mm of concat1] {};
                    \node (s2) [block, right = 1mm of s1] {};
                    
                    \node (p1) [point, right=3mm of s2] {};
                    \node (psdf) [point, above = 4mm of p1] {};
                    \node (pocc) [point, above = 14mm of psdf] {};
                    \node (pscf) [point, below = 15mm of p1] {};
                    
                    \node (sdf1) [block2, above right = 1mm and 5mm of s2.center] {};
                    \node (sdf2) [block2, right = 2mm of sdf1] {};
                    \node (sdf3) [block2, right = 2mm of sdf2] {};
                    \node (sdf4) [block2, right = 2mm of sdf3] {};
                    \node (sdf) [right = 2mm of sdf4]{\sdf};
                    
                    \node (occ1) [block1, above right = 9mm and 5mm of s2.center] {};
                    \node (occ2) [block1, right = 2mm of occ1] {};
                    \node (occ3) [block1, right = 2mm of occ2] {};
                    \node (occ4) [block1, right = 2mm of occ3] {};
                    \node (occ) [right = 2mm of occ4]{\occ};
                    
                    \node (scf1s) [block, below right = 5mm and 5mm of s2.center] {};
                    \node (scf2s) [block, right = 2mm of scf1s] {};
                    \node (scf3s) [block, right = 2mm of scf2s] {};
                    
                    \node (p2) [point, right=2.5mm of scf3s] {};
                    
                    \node (scf1) [block3, above right = 1mm and 3.5mm of scf3s.center] {};
                    \node (scf) [right = 2mm of scf1]{\escf};
                    
                    \node (sdd1) [block3, below right = 1mm and 3.5mm of scf3s.center] {};
                    \node (sdd) [right = 2mm of sdd1]{\cdd};
            
                    % descriptor
                    \node (d1) [point, right=1.7mm of s1] {};
                    \node (d2) [point, right=1.9mm of s2] {};   
                    \node (l1) [point, below=19.0mm of d2] {}; 
                    \node (d3) [point, right=2.3mm of scf1s] {};  
                    \node (l2) [point, below=10.94mm of d3] {}; 
                    \node (d4) [point, right=2.3mm of scf2s] {};  
                    \node (l3) [point, below=10.94mm of d4] {}; 
                    \node (d5) [point, right=1.8mm of scf3s] {};
                    \node (l4) [point, below=10.95mm of d5] {}; 
                    \node (concat2) [below right = 9.5mm and 2mm of d5, inner sep=0, font=\large] 
                    {$\oplus$};
                    \node (z) [right = 2mm of concat2]{$\mathbf{z}$};
            
                    % connections
                    \draw[-] (pcl) -- (encoder);
                    \draw[->] (encoder) -| (concat1.north);
                    \draw[->] (point) -| (concat1.south);
                    \draw[-] (concat1) -- (s1);
                    \draw[-] (s1) -- (s2);
                    \draw[-] (s2) -- (p1);
                    
                    \draw[->] (p1) |- (sdf1);
                    \draw[-] (sdf1) -- (sdf2);
                    \draw[-] (sdf2) -- (sdf3);
                    \draw[-] (sdf3) -- (sdf4);
                    \draw[->] (sdf4) -- (sdf);
            
                    \draw[->] (psdf) |- (occ1);
                    \draw[-] (occ1) -- (occ2);
                    \draw[-] (occ2) -- (occ3);
                    \draw[-] (occ3) -- (occ4);
                    \draw[->] (occ4) -- (occ);
                    
                    \draw[->] (p1) |- (scf1s);
                    \draw[-] (scf1s) -- (scf2s);
                    \draw[-] (scf2s) -- (scf3s);
                    
                    \draw[-] (scf3s) -- (p2);
                    \draw[-] (p2) |- (scf1);
                    \draw[->] (scf1) -- (scf);
                    
                    \draw[-] (p2) |- (sdd1);
                    \draw[->] (sdd1) -- (sdd);
            
                    \draw[-] (d1) |- (l1);
                    \draw[-] (d2) |- (l2);
                    \draw[-] (d3) |- (l3);
                    \draw[-] (d4) |- (l4);
                    \draw[->] (d5) |- (concat2);
                    \draw[->] (concat2) -- (z);
            
                    % background
                    \begin{pgfonlayer}{background}
                        % Compute a few helper coordinates
                        \path (encoder.north west)+(-0.08,0.85) node (a) {};
                        \path (z.east)+(0.4,0.3) node (b) {};
                        \path[fill=blue!15, rounded corners, draw=black, thick]
                            (a) rectangle (b);
                    \end{pgfonlayer}
                \end{tikzpicture}
                \vspace{-3pt}
                \caption{Network Structure of \mimo.}
                \label{subfig:MIMO_model_structure}
            \end{subfigure}
        } &
        \multirow{2}{*}{
            \hspace{-48pt}
            \begin{subfigure}{\linewidth}
                \centering
                \hspace{13pt}
                \begin{tikzpicture}
                    \node (sdd) at (2, 10) [inner sep=0mm]
                    {\includegraphics[width=0.50\linewidth]{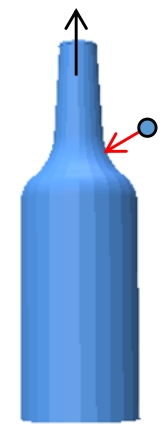}};        
                    \node (vp) [above = 15mm of sdd.center] {$\mathbf{v}_{p}$};
                    \node [below right = 3.2mm and 1mm of vp] {$\mathbf{v}_{d}$};
                \end{tikzpicture}
                \vspace{1.5pt}
                \caption{CDD.}
                \label{subfig:cdd}
            \end{subfigure}
        } &
        \multirow{2}{*}{
            \hspace{-3em}
            \begin{subfigure}{\linewidth}
                \centering
                \begin{tikzpicture}
                    \node (sim) {\includegraphics[width=0.92\textwidth]{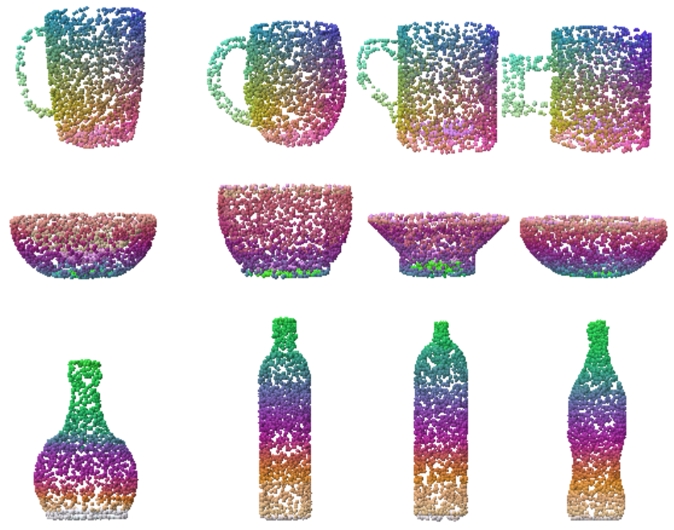}};
                    \draw[gray] (-0.9, 1.5) -- (-0.9,-1.5);
                    \node at (-1.5, -1.9) {\scalebox{0.6}{Ref. Obj.}};
                    \node at (0.6, -1.9) {\scalebox{0.6}{Categorical. Obj.}};
                \end{tikzpicture}
                \vspace{-3pt}
                \caption{Similarity Measure.}
                \label{subfig:similarity_measure}
            \end{subfigure}
        } &
        \begin{subfigure}{\linewidth}
            \hspace{55pt}
            \centering
            \includegraphics[width=1\linewidth]{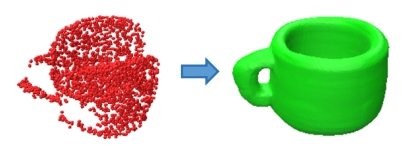}
            \caption{Shape Reconstruction.}
            \label{subfig:shape_comp}
        \end{subfigure}
        \\
        & & & 
        \begin{subfigure}{\linewidth}
            \includegraphics[width=1\linewidth]{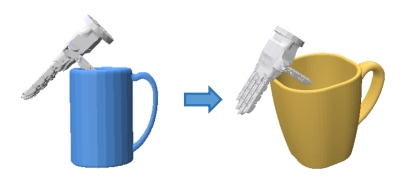}
            \caption{Grasp Transfer.}
            \label{subfig:grasp_transfer}
        \end{subfigure}
    \end{tabu}
    \caption{\mimolong (\mimo) and its applications. 
    \subref{subfig:MIMO_model_structure} \mimo takes as input an object point cloud $\mathbf{P}$ and a point coordinate $\mathbf{x}$ and outputs multiple spatial features of $\mathbf{x}$ relative to $\mathbf{P}$, including occupancy \occ, signed distance \sdf, extended space coverage feature (ESCF) \escf and closest distance direction (CDD) \cdd.
    The concatenation of activation layers of the decoder for \escf and \cdd forms the point descriptor $\mathbf{z}$ of $\mathbf{x}$. 
    \subref{subfig:cdd} The CDD is represented as the inner product of two unit vectors $\mathbf{v}_{p}$ and $\mathbf{v}_{d}$. 
    \subref{subfig:similarity_measure} The high-dimensional point descriptors of each reference object are reduced to a 3D space using Principal Component Analysis (PCA) representing the RGB channels of the color map.
    Each point of other categorical object instances is colorized according to the most similar point (smallest L1 distance in point descriptors) from the corresponding reference object.
    The \mimo can be used for \subref{subfig:shape_comp} object shape reconstruction and \subref{subfig:grasp_transfer} grasp pose transfer.}
    \label{fig:MIMO}
    \vspace{-1em}
\end{figure*}

\section{\mimo for Manipulation}
In this paper, we focus on learning task-oriented grasping and object rearrangement tasks from human demonstration videos. % in manipulation tasks.
We first introduce the \mimolong (\mimo) and its applications in \cref{sec:MIMO}, and then propose a novel grasping framework in \cref{sec:grasp_framwork} to learn and generate task-oriented grasps.

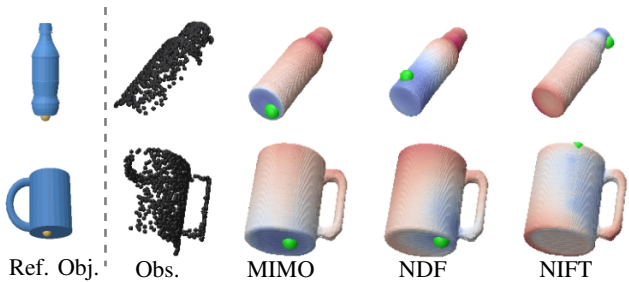
\begin{figure}[t!]
    \centering
    \begin{tikzpicture}
    \centering
        \node (pic) []
        {\includegraphics[width=0.47\textwidth] {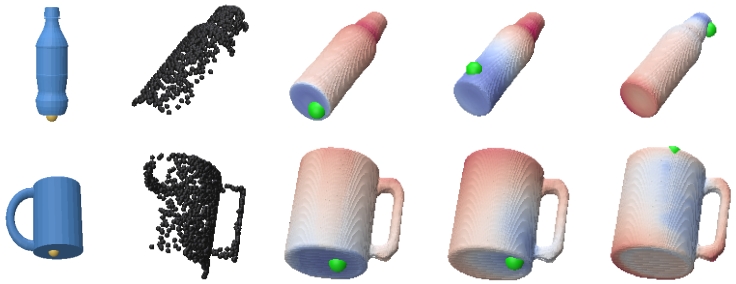}};
        \node (demo) [left=0.155\textwidth of pic.south, font=\small] {Ref. Obj.};
        \node (demo1) [left=0.095\textwidth of pic.south, font=\small] {Obs.};
        \node (demo2) [right=0.06\textwidth of demo1.center, font=\small] {\mimo};
        \node (demo3) [right=0.055\textwidth of pic.south, font=\small] {\ndf};
        \node (demo4) [right=0.16\textwidth of pic.south, font=\small] {\nift};
        \draw [dashed, color=gray, line width =1pt] (-2.8,1.75)--(-2.8,-1.7); 
    \end{tikzpicture}
    \caption{Point correspondence and shape similarity measure using point descriptors from partially-observed point clouds \mbox{(\inlinepoint{black})}. Given a point on a reference object, we colorize the novel object mesh based on the L1 distance of point descriptors to the reference point, where blue means more similar, and mark the most similar points (\inlinepoint{green}). 
    }
    \vspace{-1em}
\label{fig:sim_pts}
\end{figure}

\subsection{Multi-feature Implicit Model}
\label{sec:MIMO}
As shown in \cref{subfig:MIMO_model_structure}, \mimo uses a shared PointNet \cite{qi2017pointnet} encoder $\varepsilon(\mathbf{P})$ embedding the geometric information of the point cloud $\mathbf{P}$ in a latent code, %of dimension $k$ 
and a partly shared Multi-layer Perceptron (MLP) decoder with multiple branches, representing spatial relations of a point $\mathbf{x}$ relative to $\mathbf{P}$.
The occupancy \occ \cite{mescheder2019occupancy} and signed distance \sdf \cite{park2019deepsdf} branches enable \mimo to reconstruct object shapes.
Specifically, given the fully- or partially-observed point cloud of an object, we extract the object mesh from the trained occupancy branch using the Multi-resolution IsoSurface Extraction algorithm~\cite{mescheder2019occupancy} (see \cref{subfig:shape_comp}).
We experience that, jointly training the signed distance branch yields more precise shape reconstruction compared with training the occupancy branch alone.
Additionally, we introduce two novel feature branches, namely, 
\begin{enumerate*}
    \item the extended SCF (ESCF) branch \escf; and 
    \item the closest distance direction (CDD) branch \cdd.
\end{enumerate*}
In contrast to the SCF branch utilized in \nift~\cite{huang2023nift}, where the power spectrum of each degree in the spherical harmonics expansion is considered, our ESCF branch is directly supervised by the coefficients of spherical harmonics expansion across all orders and degrees. This enables ESCF to capture finer geometric details.
To further enhance the neural field's direction-awareness, we introduce CDD, defined as the inner product of unit vectors $\mathbf{v}_{d}$ and $\mathbf{v}_{p}$, where $\mathbf{v}_{d}$ points from a point $\mathbf{x}$ to the closest point on the object, and $\mathbf{v}_{p}$ follows a chosen principal direction, \eg pointing upward when the object is positioned vertically (see \cref{subfig:cdd}). 
Similarly to \ndf, we concatenate the activation layers of the \emph{partly-shared decoder} for \escf and \cdd as the point descriptor \eqs{$\mathbf{z} = \kappa(\mathbf{x}|\mathbf{P})$}, which forms a descriptor space to measure geometric similarity (see \cref{subfig:similarity_measure}).
Trained with four branches, our descriptor space is more informative in distinguishing fine geometric details.
In practice, we observed that the performance of the similarity measure drops when directly inferring $\mathbf{z}$ from the noisy partially-observed point cloud \eqs{$\mathbf{P}$}. 
To address this problem, we reconstruct the mesh, from which a point cloud \eqs{$\mathbf{P}_r$} is sampled as input to \mimo to infer the point descriptor \eqs{$\mathbf{z} = \kappa(\mathbf{x} \vert \mathbf{P}_r)$}.
As shown in \cref{fig:sim_pts}, \mimo finds point correspondence between the reference object and a partially-observed categorical object precisely, while \ndf yields an imprecise point correspondence and \nift often fails to distinguish the up and down direction of the bottle or the mug.
Further evaluation results are shown in \cref{subsec:MIMO_eval}.
Since all the features can be automatically computed, no further human annotation is required to collect the training dataset. 
Next, we detail the loss functions for training \mimo.

\subsubsection{Multi-task Loss Function}
\label{subsec:multi_task_loss_func}
For training \mimo, having four distinct feature branches, we combine the loss functions of each branch through a weighted sum. However, manually adjusting the weights for these loss functions is challenging. 
To address this problem, we introduce homoscedastic uncertainty~\cite{kendall2018multi} for each branch,
where the likelihood is defined as a Gaussian $p(\mathbf{y_i}|f_{\mathbf{W_i}}(\mathbf{x}))=\mathcal{N}(f_{\mathbf{W_i}}(\mathbf{x}), \sigma_i^2), i\in[1, 4]$ with the model output $f_{\mathbf{W_i}}(\mathbf{x})$ as the mean and the variance $\sigma_i$ representing the uncertainty.
The objective is to maximize the overall likelihood, or equivalently to minimize its negative log-likelihood, \ie ${\cal{L}} = \sum_{i=1}^{4}(\frac{1}{2\sigma _{i}^{2}}{\cal{L}}_{i}(\mathbf{W_{i}}) + \log(\sigma _{i}))$, 
where ${\cal{L}}_{i}$ are binary cross entropy loss for occupancy, clamped L1 loss for signed distance, and L1 losses for ESCF and CDD, respectively. 
For better numerical stability, we set $ s_{i}= \log(\sigma _{i}^{2}), i= \left \{ 1, 2, 3, 4 \right \} $ following \cite{kendall2018multi}.
Thus, the total loss is reformulated as ${\cal{L}} = \sum_{i=1}^{4}(e^{-s^{i}}{\cal{L}}_{i}(\mathbf{W_{i}})+s_{i}).$
During the training, we minimize the loss function with respect to weights of the model $\mathbf{W_{i}}$ and uncertainty $s_{i}$. 
In this way, uncertainty is automatically optimized without manual tuning.

\begin{figure*}[ht!]
    \centering
    \tikzset{
    % Specifications for style
    base/.style = {rectangle, draw=black, minimum width=24mm, minimum height=20mm, rounded corners},
    MIMO/.style = {base, fill=blue!21},
    block/.style = {rectangle, draw=black, fill=blue!15, minimum width=2mm, minimum height=6mm},
    encoder/.style = {base, minimum width=2mm, minimum height=8mm, fill=gray!20},
    gmm/.style = {base, fill=orange!60, minimum width=10mm, minimum height=8mm},
    eval/.style = {base, fill=red!20},
    b_eval/.style = {base, fill=red!50},
    sim/.style = {base, fill=green!50, minimum width=10mm, minimum height=8mm},
    }
    \tikzstyle{point}=[coordinate, on grid]
    % We need layers to draw the block diagram
    \pgfdeclarelayer{background}
    \pgfdeclarelayer{foreground}
    \pgfsetlayers{background, main, foreground}
    
    \begin{tikzpicture}
        % part 1
        \node (demo) {\includegraphics[width=0.06\textwidth] 
        {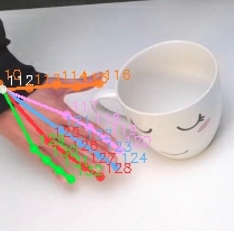}};
        \node (cap_demo) [below = 4.5mm of demo.center, align=center, font=\footnotesize] 
        {$\mathbf{P}^d, \mathbf{T}_g^d$};
        \node (demo_r) [right = 3mm of demo.east] {};
        
        \node (can) [below = 4mm of demo]
        {\includegraphics[width=0.06\textwidth] 
        {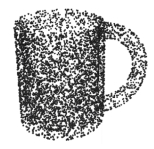}};
        \node (cap_can) [below = 4mm of can.center, align=center, font=\footnotesize] 
        {$\mathbf{P}^c$};
        \node (can_r) [right = 3mm of can.east] {};
        \node (can_1) [below = 1mm of can_r] {};
        \node (can_2) [right = 0.5mm of can.east] {};
        \node (can_3) [below = 5.3mm of can_2.center] {};
        
        \node (grasp) [above = 4mm of demo] 
        {\includegraphics[width=0.07\textwidth] 
        {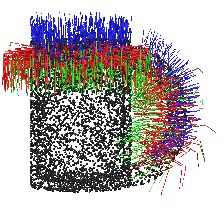}};
        \node (cap_grasp) 
        [below = 5mm of grasp.center, align=center, font=\footnotesize] 
        {$\mathbf{P}^c, \set{\mathbf{T}_g^a}$};
        \node (g_r) [right = 2mm of grasp.east] {};

        \node (MIMO_gen) [MIMO, above right = 1mm and 3mm of can_1.center] {};
        \node (mg1) [below = 1mm of MIMO_gen]{};
        \node (opt) [above = 8.5mm of mg1.center]
        {\includegraphics[width=0.07\textwidth] {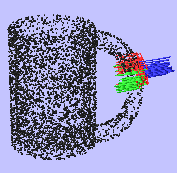}};
        \node (cap_opt) [below = 5mm of opt.center, align=center, font=\small] 
        {MIMO \\ \scalebox{0.8}{(Grasp Transfer)}};
        \node (p_g_u) [point, above = 4mm of MIMO_gen.west] {};
        \node (p_g_d) [point, below = 7.6mm of MIMO_gen.west] {};
        
        \node (MIMO_descriptor) [MIMO, above = 3mm of MIMO_gen] {};
        \node (md1) [below = 1mm of MIMO_descriptor]{};
        \node (des) [above = 7.5mm of md1]
        {\includegraphics[width=0.07\textwidth] 
        {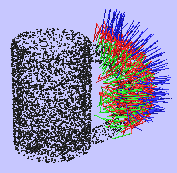}};
        \node (cap_filter) [below = 5mm of des.center, align=center, font=\small] 
        {MIMO \\ \scalebox{0.8}{(Discriminator)}};
        \node (p_des_u) [point, above = 4mm of MIMO_descriptor.west] {};
        \node (p_des_d) [point, below = 4mm of MIMO_descriptor.west] {};

        \node (p_des_r) [point, above = 6mm of MIMO_descriptor.east] {};
        \node (p_des_r2) [point, right = 3mm of p_des_r] {};

        \node (u) [right = 2mm of p_des_r, inner sep=0] {\textcircled{\raisebox{-0.2mm}{\scalebox{0.8}{$\cup$}}}};
        
        \node (sim) [sim, minimum width=16mm, minimum height=6mm, 
        right=9mm of p_des_r2, font=\small] 
        {Simulation};

        \node (gmm) [gmm, minimum width=16mm, minimum height=6mm, 
        below=6mm of sim.south, font=\small] 
        {GMM};
        
        % link
        \draw[->] (can_2.center) -- (p_g_d);

        \draw[-] (can.east) -| (can_3);
        \draw[-] (demo.east) -- (demo_r.center);
        \draw[->] (demo_r.center) |- (p_g_u);
        \draw[->] (grasp.east) -- (p_des_u);
        \draw[->] (demo_r.center) |- (p_des_d);
 
        \draw[->] (p_des_r) -- (u.west);
        \draw[->] (MIMO_gen) -| (u.south);
        \draw[->] (u.east) -- (sim.west);
        
        \node [above right = 0.1mm and 0.01mm of u.center, align=center, font=\footnotesize] 
        {$\set{\mathbf{T}_g^r}$};
        \draw[->] (sim.south) -- (gmm.north);
        \node [above right = 0.35mm and 0.1mm of gmm.north, align=center, font=\footnotesize] 
        {$\set{\bar{\mathbf{T}}_g^{r}}$};

        % part 2
        \node (pcd_i) [above right=0.5mm and 6mm of sim.east]
        {\includegraphics[width=0.05\textwidth] {figure/approaches/framework/pcd_input.png}};
        \node (cap_pcd_i) [below = 4.5mm of pcd_i.center, align=center, font=\scriptsize] {$\mathbf{P}$};
        \node (encoder) [encoder, right = 7mm of pcd_i, align=center, font=\footnotesize, minimum width=15mm]
        {$\epsilon(\mathbf{P})$};
        \node (snow) [below right=3mm and 2.2mm of encoder.north] {\scalebox{0.8}{\textcolor{blue!70}{\SnowflakeChevron}}};
        \node (set) [below = 5mm of pcd_i.center]
        {\includegraphics[width=0.06\textwidth] {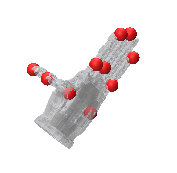}};
        \node (cap_set) [below = 3.5mm of set.center, align=center, font=\footnotesize]
        {$\mathbf{T}_g, \mathbf{P}^k$};
        
        \node (concat) [right = 2mm of encoder.east, inner sep=0, font=\large] {$\oplus$};
        \node (b1) [block, right=1mm of concat.east]{};
        \node (b2) [block, right=1mm of b1.east]{};
        \node (b3) [block, right=1mm of b2.east]{};
        \node (b4) [block, right=1mm of b3.east]{};
        \node (out) [right = 3mm of b4.east, align=center, font=\small] 
        {$\phi(\cdot)$};

        % link
        \draw[->] (pcd_i) -- (encoder);
        \draw[->] (encoder) -- (concat.west);
        \draw[->] (set) -| (concat.south);
        \draw[-] (concat) -- (b1);
        \draw[-] (b1) -- (b2);
        \draw[-] (b2) -- (b3);
        \draw[-] (b3) -- (b4);
        \draw[->] (b4) -- (out);

        % part 3
        \node (gd) [below = 14mm of gmm.south] {};
        \node (obs) [below = 12.5mm of gmm.south]
        {\includegraphics[width=0.05\textwidth] {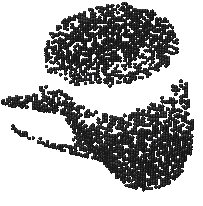}};
        \node (cap_obs) [below = 3mm of obs.center, align=center, font=\footnotesize]  
        {$\mathbf{P}^o$};

        \node (MIMO) [MIMO, right=18mm of gd.center,  minimum height=14mm, minimum width=25mm, align=center, font=\small]
        {MIMO \\ \scalebox{0.8}{(Grasp Transfer)}};
        \node (pm_u) [point, above=3mm of MIMO.west]{};
        \node (pm_d) [point, below=3mm of MIMO.west]{};
        \node (p_cap) [point, left=4mm of pm_d]{};

        \node (eval) [eval, right=6mm of MIMO.east,  minimum height=16mm, minimum width=28mm] {};
        \node (g_eval) [b_eval, above right=1mm and 8mm of MIMO.east,  minimum height=6mm, minimum width=24mm] {Evaluation};
        \node (g_ref) [b_eval, below right=1mm and 8mm of MIMO.east,  minimum height=6mm, minimum width=24mm] {Refinement};
        \node (p_el) [point, left=8mm of g_eval.south]{};
        \node (p_er) [point, right=8mm of g_eval.south]{};
        \node (p_rl) [point, left=8mm of g_ref.north]{};
        \node (p_rr) [point, right=8mm of g_ref.north]{};

        \node (g) [right=4mm of eval, font=\footnotesize] 
        {$\mathbf{T}_g^*$};
        \node [above right = 0.1mm and -2mm of obs, font=\footnotesize] 
        {$\hat{\mathbf{T}}_g$};
        \node [above right = 3mm and 37.2mm of obs.east, font=\footnotesize]
        {$\tilde{\mathbf{T}}_g$};

         % link
        \draw[->] (gmm.south) |- (pm_u);
        \draw[->] (obs) -- (pm_d);

        \draw[->] (MIMO) -- (eval);
        \draw[->] (p_el) -- (p_rl);
        \draw[->] (p_rr) -- (p_er);

        \draw[->] (eval) -- (g);

        \draw[->] (can_3.north) -| (MIMO.south);
        \draw [dashed, color=gray, line width =2pt] (13,0.8)--(13,-0.3); 
        
         % background
        \begin{pgfonlayer}{background}
            % grasp learning
            \draw [thick, rounded corners, fill=yellow, fill opacity=0.2] 
            (0.9,3.1) -- (0.9,-2.1) -- (4.4, -2.1) -- (4.4, 0.3) 
            -- (6.8, 0.3) -- (6.8, 3.1) -- cycle; 

            % inference
            \draw [thick, rounded corners, fill=yellow, fill opacity=0.2] 
            (7.1, -2.0) -- (7.1, 0.2) -- (14.6, 0.2) -- (14.6, -2.0) -- cycle;

            % grasp eval
            \draw [thick, rounded corners, fill=red, fill opacity=0.2] 
            (8.8, 3.15) -- (8.8, 0.8) -- (14.4, 0.8) -- (14.4, 3.15) -- cycle;
        \end{pgfonlayer}

        \node at (3.9, 2.77) 
        {\scalebox{0.9}{(a) Task-relevant Grasp Learning}};
        \node at (11.8, 1.1) 
        {\scalebox{0.9}{(b) Grasp Evaluation Network}};
        \node at (10.6, 0.0) 
        {\scalebox{0.9}{(c) Inference}};

    \end{tikzpicture}
    \caption{Proposed \mimo-based Grasp Framework. 
    (a) Given a human demonstration of a grasping scene, we obtain the object point cloud \eqs{$\mathbf{P}^{d}$} and a grasp pose \eqs{$\mathbf{T}_{g}^d$}. 
    We generate task-agnostic grasp poses \eqs{$\set{\mathbf{T}_g^{a}}$} using a grasp generator \cite{sundermeyer2021contact}, and use \mimo as a discriminator to select the task-relevant candidates \eqs{$\set{\mathbf{T}_g^{r}}$} based on pose descriptor similarities between \eqs{$\mathbf{T}_g^{d}$} and \eqs{$\mathbf{T}_g^{a}$}.
    Alternatively, we can directly transfer the demonstrated grasp pose \eqs{$\mathbf{T}_g^d$} to the canonical point cloud \eqs{$\mathbf{P}^c$} using \mimo. 
    We then simulate the candidates \eqs{$\set{\mathbf{T}_g^{r}}$} to find the successful grasp poses \eqs{$\set{\bar{\mathbf{T}}_g^{r}}$} to train a GMM. 
    (b) Given an object point cloud \eqs{$\mathbf{P}$}, a grasp pose \eqs{$\mathbf{T}_g$} and a set of hand keypoints \eqs{$\mathbf{P}^k$}, the grasp evaluation network encodes \eqs{$\mathbf{P}$} using the frozen encoder \eqs{$\epsilon(\cdot)$} of \mimo and outputs the grasp success probability using MLP. 
    (c) During inference, the sampled grasp pose \eqs{$\hat{\mathbf{T}}_g$} relative to the canonical point cloud \eqs{$\mathbf{P}^c$} is transferred to a partially-observed point cloud \eqs{$\mathbf{P}^o$} using \mimo, and the transferred grasp pose \eqs{$\tilde{\mathbf{T}}_g$} is evaluated and refined (if necessary) to obtain the optimal grasp pose \eqs{$\mathbf{T}_g^*$}.
    }
    \label{fig:framework}
    \vspace{-1em}
\end{figure*}

\subsubsection{Pose Descriptor}

\label{subsec:g_pose_sim}
Similarly to~\cite{simeonov2022neural}, we adopt the Basis Point Set (BPS) \cite{prokudin2019efficient} sampling strategy, and concatenate the point descriptors of a set of points around an object as their pose descriptor \eqs{$\mathbf{Z}$}.
Specifically, given a set of points \eqs{$\mathbf{X} \in \mathbb{R}^{N\times 3}$} sampled from a rigid object \objB in pose $\mathbf{T}$ around the point cloud \eqs{$\pclA$} of object \objA, we obtain pose descriptor of \objB using the trained \mimo of object category A, \ie \eqs{$\dposeAB = \varphi(\mathbf{T}, \mathbf{X} \vert \pclA)$}. 
It measures the similarity of the poses relative to \objA, where similar poses have a small L1 distance between their pose descriptors.
Speaking in terms of the example in \cref{fig:teasor}, \objA would be an instance of the ``mug'' class, while \objB would be the hand and, therefore, \poseAB associated to a grasp pose $\mathbf{T}$.
Similarly to \cref{sec:MIMO},
% To address this problem, 
we reconstruct the mesh of \objA, from which a point cloud \eqs{$\pclAr$} is sampled as input to \mimo to infer the pose descriptor \eqs{$\dposeAB = \varphi(\mathbf{T}, \mathbf{X} \vert \pclAr)$}.
\subsubsection{Pose Transfer}
\label{subsec:g_pose_trans}

Given a trained \mimo of object category A, a reference pose descriptor \eqs{\poseABref} and a pair of arbitrary object instances (\objAnew,\objBnew) from category A and B, we optimize the pose of \objBnew relative to \objAnew by \eqs{$\mathbf{T}^* = \argmin_{\mathbf{T}}\Vert \varphi(\mathbf{T}, \mathbf{X}\vert \pclAnew) - \dposeABref\Vert_1$}, where \eqs{$\pclAnew$} is the reconstructed point cloud of \objAnew.
We adopt the same optimization procedure as in~\cite{simeonov2022neural}. 
In a visual imitation learning (VIL) setup, the reference pose descriptor can be derived from human demonstration videos.
Specifically, we find the closest point pair on \objA and \objB at the last timestep of the demonstration as keypoints.
Similarly to~\cite{simeonov2023se}, we then sample BPS around keypoints of \objA and \objB respectively, to compute the corresponding reference pose descriptors.
which can be used to transfer \objAnew and \objBnew to align with \objA and \objB, respectively, with the optimization steps described above. 
The rearrangement target pose of \objBnew relative to \objAnew can be derived from the optimized poses.
Note that the keypoints and sampled BPS do not need to lie on the object.
We refer interested readers to~\cite{simeonov2023se} for more details.
In terms of grasping, where \objB is the human or robot hand and \objA is an arbitrary object to be grasped, the pose descriptors measure the grasp similarity, which can be used for transferring grasps to similar objects. Next, we introduce a novel grasp framework based on \mimo.

\subsection{\mimo-based Grasp Framework}
\label{sec:grasp_framwork}

Leveraging \mimo's strengths in measuring pose similarities and transferring poses, we introduce a framework designed to learn task-specific grasping and object rearrangement from one or multiple human demonstrations. This framework can generate optimal grasp poses for new object instances based on partial observations, as shown in \cref{fig:framework}.

\subsubsection{Human Observation}
\label{subsec:human_obs}
Given human demonstration videos consisting of sequences of RGB and depth images of a manipulation task,
we estimate the hand poses in all frames using \cite{lin_mesh_2021} and train a movement primitive~\cite{Zhou2019} representing the hand motion.
We then determine the grasping timestep $t_g$ and detect grasp pose \eqs{$\mathbf{T}_g\in SE(3)$} following \cite{gao_bikvil_2024}.
The object being grasped is the source object \objs, and the other object, which sets a reference frame for placing \objs at the last timestep $t_T$, is the target object \objt.
We obtain the segmented point clouds of both objects at $t_g$ and $t_T$ using Grounded SAM \cite{liu2023grounding,kirillov_segment_2023}.

\subsubsection{Task-oriented Grasp Learning}
\label{subsec:grasp_learning}

As shown in \cref{fig:framework} (a), we generate task-agnostic grasp candidates \eqs{$\set{\mathbf{T}_g^a}$} using \cite{sundermeyer2021contact} on a canonical point cloud \eqs{$\mathbf{P}^c$} for the class of the source object \objs.
We present two strategies to obtain task-relevant grasp candidates, \ie \begin{enumerate*}[label=(\roman*)]
    \item using \mimo as a discriminator for pose similarity to find the most similar grasps in \eqs{$\set{\mathbf{T}_g^a}$ to $\mathbf{T}_g^d$} (see \cref{subsec:g_pose_sim}); or 
    \item using \mimo to directly transfer the demonstrated grasp \eqs{$\mathbf{T}_g^d$} relative to \eqs{$\mathbf{P}^d$} to a set of candidate grasps relative to canonical space (see \cref{subsec:g_pose_trans}). 
\end{enumerate*}
We fuse the task-relevant grasp candidates \eqs{$\set{\mathbf{T}_g^r}$} from the two strategies and simulate them with a humanoid hand in Issac Gym \cite{makoviychuk2021isaac}.
Specifically, the grasp is successful if the object is picked up and does not drop after a random shaking action. 
We then simulate the object rearrangement given the successful grasps and determine the set of task-relevant grasps if the tasks are accomplished without failure (see \cref{sec:experiments} for a definition of possible tasks). 
The successful and task-relevant grasps \eqs{$\set{\bar{\mathbf{T}}_g^r}$} in canonical space are used to train a GMM on a Riemannian manifold (\ie $\mathbb{R}^3 \times \mathcal{S}^3$), which can be used to generate task-oriented grasps.

\subsubsection{Grasp Evaluation}
\label{subsec:grasp_eval_net}

The sampled task-relevant grasps from the GMM are not guaranteed to be successful. 
To address this problem, we propose a \emph{task-agnostic} grasp evaluation network to compute the success probability of a grasp pose \eqs{$\mathbf{T}_g$} relative to an arbitrary point cloud \eqs{$\mathbf{P}$} (see \cref{fig:framework}).
We first encode \eqs{$\mathbf{P}$} using the frozen encoder of \mimo, \ie \eqs{$\mathbf{c} = \varepsilon(\mathbf{P})$}. 
We then use a MLP decoder conditioned on $\mathbf{c}$ to predict the success probability given a set of keypoints \eqs{$\mathbf{P}^k$} on the humanoid hand representing its pose, \ie \eqs{$\phi (\mathbf{T}_g, \mathbf{P}^k|\varepsilon (\mathbf{P})) \in [0, 1]$}.
We train this model using a binary cross-entropy loss on a dataset fusing the task-agnostic grasp candidates for all objects in all tasks, along with their binary labels indicating successful grasps obtained in \cref{subsec:grasp_learning}.

\subsubsection{Inference}\label{subsec:grasp_inference}
During inference, we sample grasp poses \eqs{$\hat{\mathbf{T}}_g$} from the trained GMM relative to the canonical point cloud \eqs{$\mathbf{P}^c$}, which are then transferred to a partially-observed point cloud \eqs{$\mathbf{P}^o$} of a novel categorical instance following \cref{subsec:g_pose_trans}.
We compute the success probability of the transferred grasps \eqs{$\tilde{\mathbf{T}}_g$} using the trained task-agnostic grasp evaluation network.
If the grasp success probability is lower than a certain threshold $\xi$, we refine the grasp pose by maximizing the grasping success likelihood using the grasp evaluation network from \cref{subsec:grasp_eval_net}, \ie \eqs{$\Delta \mathbf{T}_g^* = \argmax_{\Delta \mathbf{T}_g}\phi (\Delta \mathbf{T}_g\mathbf{T}_g, \mathbf{P}^k|\varepsilon (\mathbf{P}))$}, 
and finally obtain the optimal grasp pose \eqs{$\mathbf{T}_g^* = \Delta \mathbf{T}_g^*\mathbf{T}_g$}.

\section{Evaluation}
\label{sec:experiments}
We evaluate the proposed \mimo and grasping framework in different manipulation tasks and compare with \ndf \cite{simeonov2022neural}, \rndf \cite{simeonov2023se}, and \nift \cite{huang2023nift}. 
More details, evaluation videos, and source code are available at \href{https://sites.google.com/view/mimo4}{https://sites.google.com/view/mimo4}.

\subsection{Evaluation of \mimo in Simulation}\label{subsec:MIMO_eval}
We first evaluate the performance of \mimo against different approaches from the state of the art.
To show the effectiveness of 
the novel ESCF and CDD features in \mimo (denoted \mimoIV), we provide additional evaluation results of a variant of \mimo (denoted \mimoIII) with three branches in the decoder to predict occupancy, signed distance, and SCF separately.

\subsubsection{Generation of Training Data}
Training \mimo can be done without manual annotation of the training data. \ndf and \nift each provide their own datasets that could be used for training. However, we observed two issues in these datasets, namely 
\begin{enumerate*}[label=(\roman*)]
    \item the bottom of the bottle's meshes from \ndf is hollowed out, which influences the shape reconstruction quality;
    \item the mesh scaling is non-uniform, leading to wrong labels for SCF and signed distance.

\end{enumerate*}
Therefore, we generate a new dataset made from watertight meshes from the ShapeNet dataset~\cite{chang2015shapenet} using~\cite{stutz2018learning} with rendered point clouds for each mesh.
The remainder of the data generation and training of the models is similar to the procedure used for \nift. 
We train \nift and our model using the new dataset on a single NVIDIA A100 GPU, and use the pre-trained weights of \ndf and \rndf provided by the authors. 

\subsubsection{Setup and Metrics}
We consider three settings, namely \begin{enumerate*}[label=(S\arabic*)]
    \item \label{item:demo10}10 demonstrations and four viewpoints, where the point cloud is fused from 4 depth cameras at 4 corners of the table;
    \item \label{item:demo1_4}a single demonstration and four viewpoints, with the same camera positions as before; and
    \item \label{item:demo1}a single demonstration and single viewpoint, in which the mug handle and bottle opening are visible. 
\end{enumerate*}
We use BPS for all models in the evaluation tasks.
To evaluate SE(3)-equivariance of the trained neural fields, we distinguish between upright (\upPose) and arbitrary (\arbiPose) initial object poses, where the objects are positioned upright on the table for \upPose while the objects are arbitrarily positioned in the air for \arbiPose.
For \mimoIV and \mimoIII, we reconstruct object shapes from partial observations and transfer poses as discussed in \cref{subsec:g_pose_trans}. 
The overall task is successful if the object is grasped without dropping (grasp success) and the bowl/bottle stands upright on a shelf, or the mug is hung on the rack without penetration at the optimized target pose (placement success).

\subsubsection{Comparison with \ndf} \label{sec:ndf_scenarios}
We use the simulation environment and evaluation proposed from \ndf, including 3 pick-and-place tasks: \begin{enumerate*}[label=(T\arabic*)]
    \item \label{item:t1}picking a mug by the rim and placing it on the rack by the handle; 
    \item \label{item:t2}picking a bowl and placing it on the shelf; and 
    \item \label{item:t3}picking a bottle from the side and placing it on the shelf. 
\end{enumerate*}
We conduct 100 trials for each task under the two settings \ref{item:demo10} and \ref{item:demo1}% and both sampling strategies (BPS/IBS)
, and upright and arbitrary object poses respectively.

\begin{figure}[t!]
    \centering
    \hspace{-1em}
    \begin{tikzpicture}
        \node (image) at (0.0, 0) {\includegraphics[width=0.97\columnwidth]{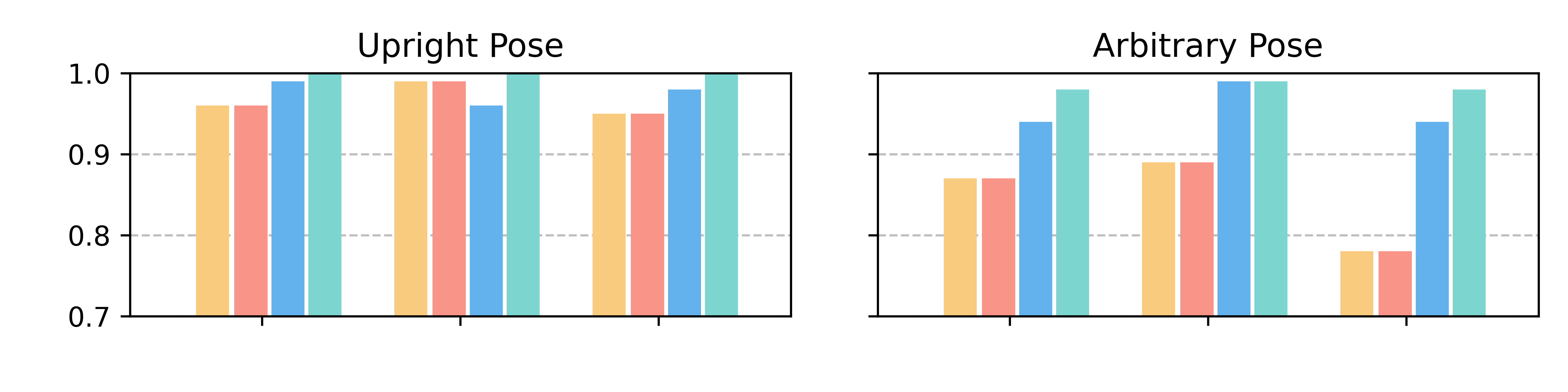}
        };
        \node at (-4.2, 0.0) {
            \scalebox{0.7}{
                \rotatebox{90}{Success Rate}
            }
        };
        \node at (-2.8, -0.9) {\scalebox{0.7}{Grasp}};
        \node at (-1.7, -0.9) {\scalebox{0.7}{Place}};
        \node at (-0.7, -0.9) {\scalebox{0.7}{Overall}};
        \node at ( 1.2, -0.9) {\scalebox{0.7}{Grasp}};
        \node at ( 2.3, -0.9) {\scalebox{0.7}{Place}};
        \node at ( 3.3, -0.9) {\scalebox{0.7}{Overall}}; 
    \end{tikzpicture}
    \vspace{-2em}  
    \caption{Success rate of the pick-and-place tasks \ref{item:t1}-\ref{item:t3} with unseen objects under setting \ref{item:demo10} for models \ndf\raisebox{-1.0mm}{\sampleline{c_ndf,line width=2mm}}, \nift\raisebox{-1.0mm}{\sampleline{c_nift,line width=2mm}}, \mimoIII\raisebox{-1.0mm}{\sampleline{c_mimo3,line width=2mm}}, and \mimoIV\raisebox{-1.0mm}{\sampleline{c_mimo4s,line width=2mm}}, respectively.}
    \label{fig:success_rate_demo10}
    \vspace{-1em}
\end{figure}

\begin{figure}[t!]
    \centering
    \hspace{-1em}
    \begin{tikzpicture}
        \node (image) at (0.1, 0) {\includegraphics[width=0.95\columnwidth]{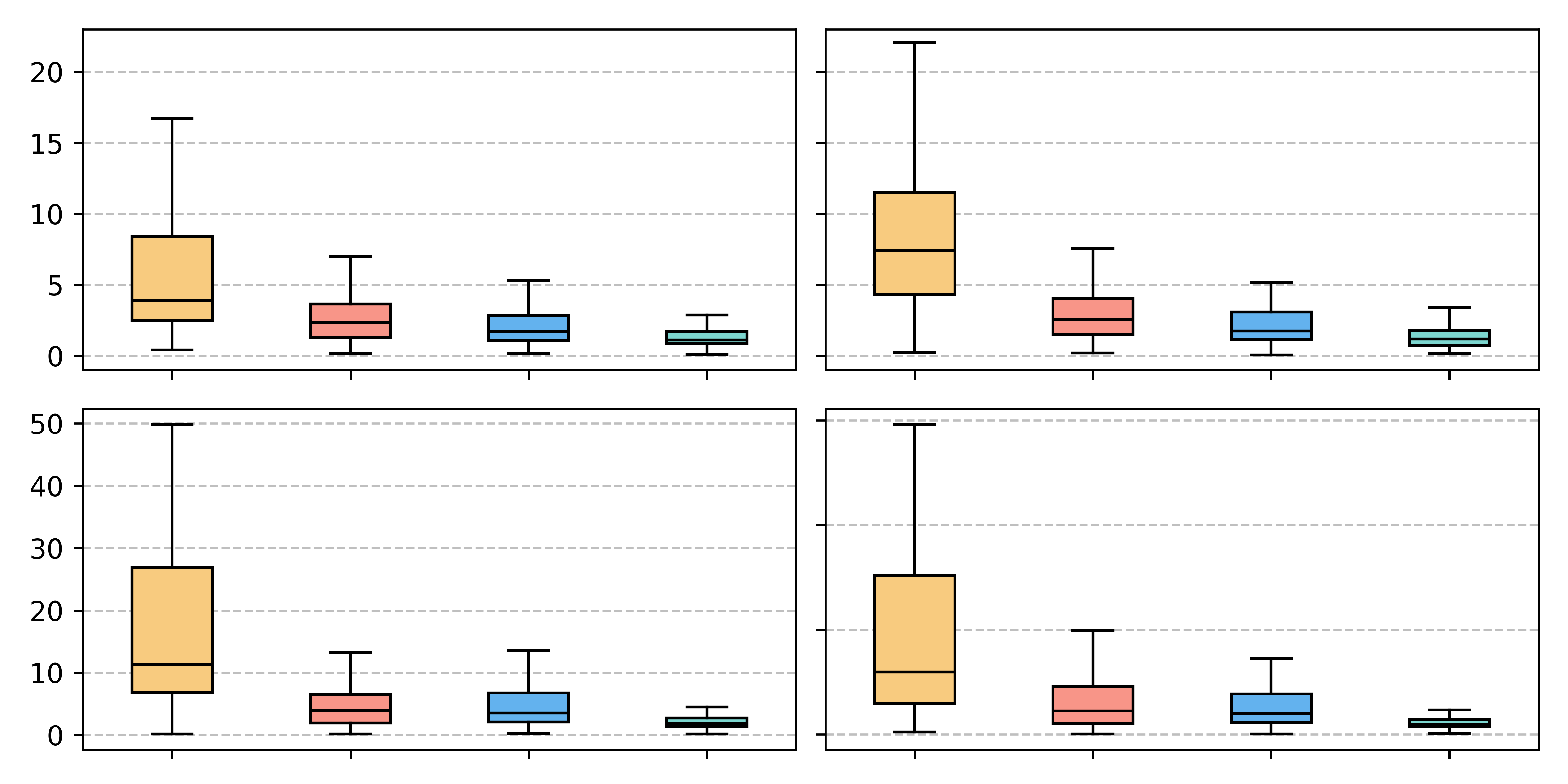}
        };
        \node at (-4.2, 1) {
            \scalebox{0.7}{
                \rotatebox{90}{Angle Error (\unit{\degree})}
            }
        };
        \node at (-4.2, -0.9) {
            \scalebox{0.7}{
                \rotatebox{90}{Angle Error (\unit{\degree})}
            }
        };
        \node[fill=white] at (-1.0,  1.6) {\scalebox{0.7}{\ref{item:demo10}, Upr. Pose \upPose}};
        \node[fill=white] at ( 2.9,  1.6) {\scalebox{0.7}{\ref{item:demo10}, Arb. Pose \arbiPose}};
        \node[fill=white] at (-1.0, -0.4) {\scalebox{0.7}{\ref{item:demo1}, Upr. Pose \upPose}};
        \node[fill=white] at ( 2.9, -0.4) {\scalebox{0.7}{\ref{item:demo1}, Arb. Pose \arbiPose}};
        \node at (-3.1, -2.1) {\scalebox{0.7}{\ndf}};
        \node at (-2.2, -2.1) {\scalebox{0.7}{\nift}};
        \node at (-1.25, -2.1) {\scalebox{0.7}{\mimoIII}};
        \node at (-0.3, -2.1) {\scalebox{0.7}{\mimoIV}};
        \node at (0.75, -2.1) {\scalebox{0.7}{\ndf}};
        \node at (1.65, -2.1) {\scalebox{0.7}{\nift}};
        \node at (2.65, -2.1) {\scalebox{0.7}{\mimoIII}};
        \node at (3.60, -2.1) {\scalebox{0.7}{\mimoIV}};
    \end{tikzpicture}
    \caption{Angle error of bowls and bottles. Colors as \cref{fig:success_rate_demo10}.}
    \label{fig:angle_dev}
    \vspace{-1.5em}
\end{figure}
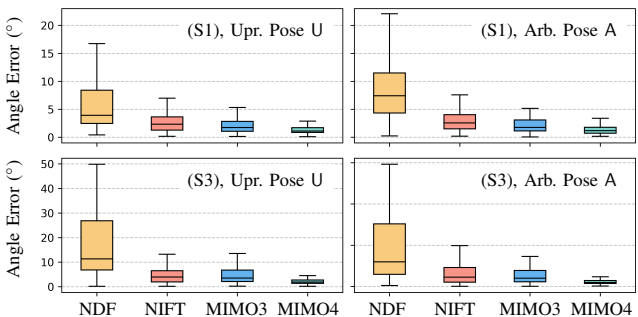

\begin{table*}[!t]
\caption{Unseen object pick-and-place success rate with setting \ref{item:demo1} (single viewpoint, single demonstration).}
\centering                  %ccccccccccccccc
\begin{tabularx}{\textwidth}{clXXXXXXXXXXXX}  
    \toprule
             &         & \multicolumn{3}{c}{Mug \ref{item:t1}}  & \multicolumn{3}{c}{Bowl \ref{item:t2}}  & \multicolumn{3}{c}{Bottle \ref{item:t3}}  & \multicolumn{3}{c}{Mean} \\
             \cmidrule(lr){3-5}\cmidrule(lr){6-8}\cmidrule(lr){9-11}\cmidrule(lr){12-14}
             &        & Grasp     & Place     & Overall   & Grasp     & Place     & Overall   & Grasp     & Place     & Overall   & Grasp     & Place     & Overall   \\ 
    \midrule
    & \ndf     & 0.95      & 0.73      & 0.72      & 0.89      & 0.93      & 0.84      & 0.90      & 0.69      & 0.65      & 0.91      & 0.78      & 0.74      \\ 
    & \rndf   & 0.90      & 0.77      & 0.69      & 0.90      & \zz{1.00}      & 0.90      & 0.53      & \zz{0.97}      & 0.51      & 0.78      & 0.91      & 0.70      \\ 
    & \nift    & 0.99      & 0.92      & 0.92      & 0.98      & \zz{1.00}      & 0.98      & 0.96      & 0.94      & 0.90      & 0.98      & 0.95      & 0.93      \\
    & \mimoIII   & \zz{1.00}      & 0.92      & 0.92      & 0.99      & \zz{1.00}      & \zz{0.99}       & 0.92      & 0.93      & 0.91      & 0.97      & 0.95      & 0.94      \\
    \multirow{-5}{*}{\rotatebox[origin=c]{90}{\scalebox{0.9}{Upr. Pose \upPose}}}& \mimoIV & \zz{1.00} & \zz{0.98} & \zz{0.98} & \zz{1.00}      & 0.99      & \zz{0.99}      & \zz{0.97}      & \zz{0.97}      & \zz{0.95}      & \zz{0.99} & \zz{0.98} & \zz{0.97} \\ \hline
    & \ndf     & 0.53      & 0.58      & 0.34      & 0.76      & 0.80      & 0.64      & 0.42      & 0.91      & 0.40      & 0.57      & 0.76      & 0.46      \\
    & \rndf   & 0.50      & 0.70      & 0.35      & 0.78      & 0.97      & 0.77      & 0.12      & 0.90      & 0.09      & 0.47      & 0.86      & 0.40      \\
    & \nift    & 0.46      & 0.90      & 0.42      & 0.96      & 0.96      & 0.94      & 0.38      & 0.93      & 0.37      & 0.60      & 0.93      & 0.58      \\
    & \mimoIII   & 0.86      & 0.94      & 0.80      & 0.94      & \zz{0.99}      & 0.94      & 0.77      & 0.87      & 0.71      & 0.86      & 0.93      & 0.82      \\
    \multirow{-5}{*}{\rotatebox[origin=c]{90}{\scalebox{0.9}{Arb. Pose \arbiPose}}} & \mimoIV & \zz{0.92} & \zz{0.97} & \zz{0.90} & \zz{0.98}      & 0.97      & \zz{0.95}      & \zz{0.95} & \zz{0.97} & \zz{0.93} & \zz{0.95} & \zz{0.97} & \zz{0.93} \\
    \bottomrule
\end{tabularx}
\label{tab:ndf2}
\vspace{-1.3em}
\end{table*}

As shown in \cref{fig:success_rate_demo10}, all approaches achieve high success rates of tasks \ref{item:t1}-\ref{item:t3} for the setting \ref{item:demo10}. 
We observe that \mimoIV achieves the best result in all cases, with \mimoIII ranked slightly below. 
It is important to note that the overall success rates of \mimoIV only drop by $2\%$ in arbitrary pose compared to upright pose, which is less than other approaches, showcasing a better SE(3)-equivariance property of our neural descriptor field.
In contrast, as shown in \cref{tab:ndf2}, \mimoIV significantly outperforms others in setting \ref{item:demo1}, especially in the case of arbitrary object poses in tasks \ref{item:t1} and \ref{item:t2}.
We highlight the best success rates of each (sub-)task. 
\mimoIII, \nift and \rndf only perform slightly or equally well than \mimoIV in the placing phase of \ref{item:t2}, where bowls are involved. 
The reason is that the partially-observed point cloud of bowls with large opening has already covered a large portion of the object and it is much easier to distinguish the up and down direction compared to the mugs and bottles used in \ref{item:t1} and \ref{item:t2}. 
As discussed in \cref{sec:MIMO} and \cref{fig:sim_pts}, \ndf and \nift often fail to distinguish the top and bottom of the bottle and mug handle.
Inaccurate correspondences can cause objects to be transformed into wrong poses, leading to low success rates in \ref{item:t1} and \ref{item:t2}.
In contrast, our descriptor field is more informative, achieving more accurate pose transfer and thus higher success rates.
We showcase the pose accuracy in \cref{fig:angle_dev} by computing the angle error between the object's upright direction and the gravity direction at the target pose for bowls and bottles in \ref{item:t2} and \ref{item:t3}. 
A smaller angle indicates a more precise placement pose.
We observe that our \mimoIV has the smallest average angle error and smallest variance across all tasks, further verifying the superiority of our neural descriptor.

\begin{table}[!t]
    \caption{Success rates of unseen object rearrangement. \upPose and \arbiPose stand for upright and arbitrary poses, respectively.}
    \centering
    % \resizebox{\columnwidth}{18mm}{
    \begin{tabularx}{\columnwidth}{M{0.03\columnwidth}M{0.14\columnwidth}M{0.04\columnwidth}M{0.04\columnwidth}M{0.04\columnwidth}M{0.04\columnwidth}M{0.04\columnwidth}M{0.04\columnwidth}M{0.04\columnwidth}M{0.04\columnwidth}}
        \toprule
        & & \multicolumn{2}{c}{\ref{item:t4}} & \multicolumn{2}{c}{\ref{item:t5}} & \multicolumn{2}{c}{\ref{item:t6}} & \multicolumn{2}{c}{Mean}       \\ 
        \cmidrule(lr){3-4}\cmidrule(lr){5-6}\cmidrule(lr){7-8}\cmidrule(lr){9-10}
        & Models  & \upPose   & \arbiPose   & \upPose   & \arbiPose   & \upPose   & \arbiPose  & \upPose  & \arbiPose  \\ \arrayrulecolor{black!80}\hline
        \multirow{3}{*}{\ref{item:demo10}}
        & \rndf   & 0.71           & 0.55           & 0.75           & 0.75           & 0.80           & 0.54          & 0.75          & 0.61          \\
        & \mimoIII   & \textbf{0.91}  & \textbf{0.87}  & \textbf{0.92}  & \textbf{0.91}  & 0.84           & 0.85          & \textbf{0.89} & 0.88          \\
        & \mimoIV & 0.88           & 0.85           & 0.91           & 0.89           & \textbf{0.87}  & \textbf{0.93} & \textbf{0.89} & \textbf{0.89} \\ \arrayrulecolor{gray!40}\hline
        \multirow{3}{*}{\ref{item:demo1_4}}
        & \rndf   & 0.56           & 0.53           & 0.64           & 0.61           & 0.12           & 0.18          & 0.44          & 0.44          \\
        & \mimoIII   & 0.89           & 0.89           & \textbf{0.90}  & \textbf{0.88}  & 0.85           & 0.87          & 0.88          & 0.88          \\
        & \mimoIV & \textbf{0.92}  & \textbf{0.92}  & \textbf{0.90}  & 0.87           & \textbf{0.91}  & \textbf{0.93} & \textbf{0.91} & \textbf{0.92} \\ \hline
        \multirow{3}{*}{\ref{item:demo1}}
        & \rndf   & 0.29           & 0.21           & 0.10           & 0.13           & 0.16           & 0.07          & 0.18          & 0.14          \\
        & \mimoIII   & 0.85           & 0.85           & 0.88           & 0.87           & 0.72           & 0.70          & 0.82          & 0.81          \\
        & \mimoIV & \textbf{0.89}  & \textbf{0.86}  & \textbf{0.90}  & \textbf{0.88}  & \textbf{0.90}  & \textbf{0.83} & \textbf{0.90} & \textbf{0.86} \\
        \arrayrulecolor{black}\bottomrule
    \end{tabularx}
    \label{tab:rndf}
    \vspace{-1em}
\end{table}

\subsubsection{Comparison with \rndf}

We adopt the simulation environments from \rndf \cite{simeonov2023se} with three tasks, namely: \begin{enumerate*}[label=(T\arabic*), start=4]
    \item \label{item:t4}hanging a mug on the hook of a rack; 
    \item \label{item:t5}placing a bowl on a mug; and
    \item \label{item:t6}and placing a bottle in a container. 
\end{enumerate*}
All three settings are considered, namely \ref{item:demo10}, \ref{item:demo1_4} and \ref{item:demo1}.
In contrast to the experiments in \cref{sec:ndf_scenarios}, we focus only on the target configurations of the object and neglect the grasp procedure for this evaluation. 
The task is successful if the source object is placed on the target object without falling or exerting a large interaction force. 
We conduct 100 trials for each task and compute the success rates.
As shown in \cref{tab:rndf}, \mimoIV and \mimoIII perform equally well in setting \ref{item:demo10} with a success rate of about 89\%. 
In both settings \ref{item:demo1_4} and \ref{item:demo1}, \mimoIV significantly outperforms \rndf by about 48\% and 70\%, respectively.
Therefore, we do not need the extra alignment and refinement steps as in \nift~\cite{simeonov2023se}.
Note that \mimoIII's performance drops in \ref{item:demo1_4} and further in \ref{item:demo1}, showcasing the effectiveness of the novel ESCF and CDD feature in the partly shared decoder of \mimo.

\subsection{Evaluation of the Grasping Framework}
To evaluate the performance of \mimo in the context of our grasping framework presented in \cref{sec:grasp_framwork}, we performed multiple experiments in simulation using Isaac Gym~\cite{makoviychuk2021isaac} and on the humanoid robots \armarVI~\cite{Asfour2019} and \armarDE in real-world manipulation tasks.
We define four tasks, namely: 
\begin{enumerate*}[label=(T\arabic*), start=7] 
    \item \label{item:t7}grasp a mug at its rim and place it upright in a container;
    \item \label{item:t8}grasp a mug at its handle and pour into a bowl;
    \item \label{item:t9}grasp a bottle at its neck and place it upright in a container; and
    \item \label{item:t10}grasp a bottle at its body and pour it into a bowl. 
\end{enumerate*}
Object poses are randomly initialized, with mugs positioned to ensure handle visibility.
We use \mimoIV to reconstruct object shapes from the partially-observed object point clouds.
The grasp poses are sampled from the GMM, transferred to the observed objects, and evaluated by the grasp evaluator (see \cref{subsec:grasp_eval_net}). 
If the estimated success probability drops below $0.9$, the grasp pose is optimized with a learning rate of $10^{-3}$ (see \ref{subsec:grasp_inference}). 
We then optimize the target pose for rearrangement using \mimoIV and execute the grasp and rearrangement action.

\subsubsection{Evaluation in Simulation}
\label{subsec:eval_grasp_sim}
We simulate a humanoid hand in Issac Gym equipped with a depth camera positioned in front of a table.
We use \nift with BPS as a baseline approach without the grasp evaluation and refinement.
Note that tasks \ref{item:t7}-\ref{item:t10} are successful if both grasping and rearrangement are successful.
We execute each task for 50 trials under setting \ref{item:demo1}. 
As shown in \cref{tab:isaac}, \mimoIV outperforms \nift in all tasks, especially in task \ref{item:t9} by about 72\%. \nift cannot differentiate between the top and bottom of the bottle and, therefore, fails to place the bottle in the container. 
In contrast, \mimoIV achieves higher success rates, benefiting from the reconstructed shape and the powerful descriptor space. 
In addition, \mimoIV achieves an average success rate of 95\% for grasping, including difficult side grasps at the mug handle, which demonstrates the effectiveness of our grasp evaluator.

\subsubsection{Evaluation in the Real World}
\begin{figure}
    \centering
    \begin{subfigure}{0.47\linewidth}
        \begin{minipage}[b]{1\textwidth}
        \centering
        \includegraphics[width=1\textwidth]{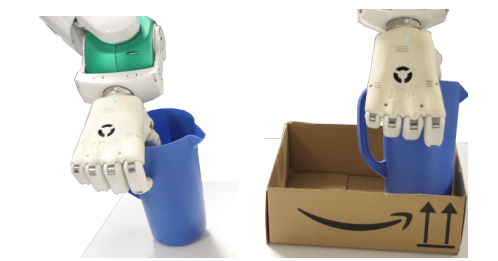} 
        \end{minipage}
        \caption{Mug Pick and Place \ref{item:t7}.}
        \label{fig:mug_place}
    \end{subfigure}
    \hspace{0.4em}
    \begin{subfigure}{0.47\linewidth}
        \begin{minipage}[b]{1\textwidth}
        \centering
        \includegraphics[width=1\textwidth]{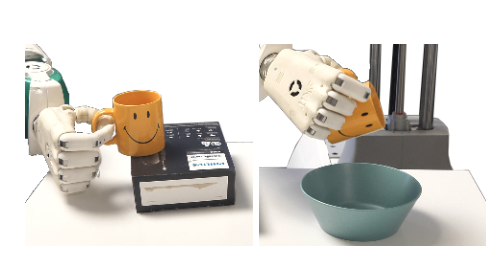} 
        \end{minipage}
        \caption{Mug Pick and Pour \ref{item:t8}.}
        \label{fig:mug_pour}
    \end{subfigure}
    \vspace{0.8em}
    \begin{subfigure}{0.47\linewidth}
        \begin{minipage}[b]{1\textwidth}
        \centering
        \includegraphics[width=1\textwidth]
        {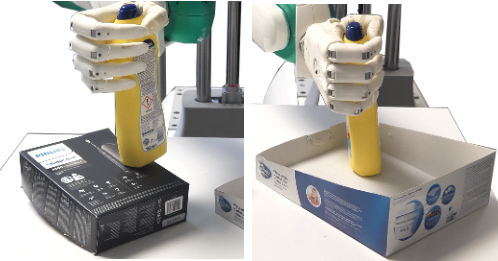} 
        \end{minipage}
        \caption{Bottle Pick and Place \ref{item:t9}.}
        \label{fig:bottle_place}
    \end{subfigure}
    \hspace{0.4em}
    \begin{subfigure}{0.47\linewidth}
        \begin{minipage}[b]{1\textwidth}
        \centering
        \includegraphics[width=1\textwidth]{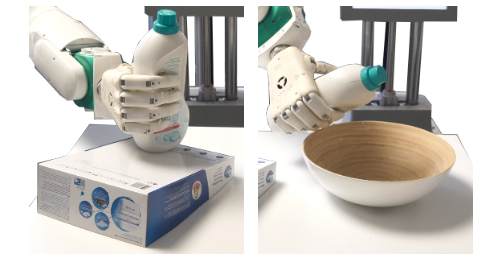} 
        \end{minipage}
        \caption{Bottle Pick and Pour \ref{item:t10}.}
        \label{fig:bottle_pour}
    \end{subfigure}
    \vspace{-1em}
    \caption{Real-world experiments on \armarDE.}
    \label{fig:real_exp}
    \vspace{-1.8em}
\end{figure}

Similarly to \cref{subsec:eval_grasp_sim}, we replicate tasks \ref{item:t7}-\ref{item:t10} using the same GMM and \mimoIV with two humanoid robots: \armarDE and \armarVI. 
We use an Azure Kinect camera mounted on the robot head to obtain RGB and depth images and extract object point clouds as explained in \cref{subsec:human_obs}. 
For the experiments on \armarDE, 
the grasp pose was validated and executed using the
mobile manipulation framework~\cite{pohl2024memorycentered}.
On \armarVI, we use a task-space impedance controller to execute the motions generated by the learned movement primitives (see \cref{subsec:human_obs}), where the target poses are the corresponding grasp pose in the grasp phase and the object rearrangement pose in the placement or pouring phase. 
We show qualitative results in \cref{fig:real_exp} and in the accompanying video, showcasing the efficacy of our approach in one-shot imitation learning of manipulation tasks.

\begin{table}[!t]
    \caption{The success rates of unseen object grasping ($\mathsf{G}$) and rearrangement ($\mathsf{R}$).}
    \centering
    \begin{tabularx}{\columnwidth}{cXXXXXXXXXX}
        \toprule
        & \multicolumn{2}{c}{\ref{item:t7}} & \multicolumn{2}{c}{\ref{item:t8}} & \multicolumn{2}{c}{\ref{item:t9}} & \multicolumn{2}{c}{\ref{item:t10}} & \multicolumn{2}{c}{Mean}       \\ 
        \cmidrule(lr){2-3}\cmidrule(lr){4-5}\cmidrule(lr){6-7}\cmidrule(lr){8-9}\cmidrule(lr){10-11}
        Models  & $\mathsf{G}$   & $\mathsf{R}$   & $\mathsf{G}$   & $\mathsf{R}$   & $\mathsf{G}$   & $\mathsf{R}$  & $\mathsf{G}$  & $\mathsf{R}$  
        & $\mathsf{G}$  & $\mathsf{R}$ \\ 
        \arrayrulecolor{black!80}\hline
        \multirow{2}{*}{}
        \nift   & 0.80         & 0.62           & 0.92           & 0.80       
              & 0.86        & 0.08         & 0.92           & 0.68    
              & 0.88    & 0.55      \\
        \mimoIV   & \textbf{0.94}   & \textbf{0.88}  & \textbf{0.96}   & \textbf{0.94}      & \textbf{0.90}        & \textbf{0.80}      & \textbf{0.98}        & \textbf{0.88}
        & \textbf{0.95} & \textbf{0.88}   \\
        \arrayrulecolor{black}\bottomrule
    \end{tabularx}
    \label{tab:isaac}
    \vspace{-1em}
\end{table}

%===============================================================================

\section{Conclusion}
\label{sec:conclusion}
We propose \mimolong (\mimo), a novel implicit neural field that provides informative and SE(3)-equivariant point and pose descriptors for shape similarity measure. 
Trained on multiple spatial features, \mimo facilitates finer correspondence detection and more accurate pose transfer compared to state-of-the-art approaches. 
\mimo also allows for shape reconstruction to account for partial observations.
Based on \mimo, we propose a task-oriented grasping and object rearrangement framework with a novel evaluation and refinement procedure to further increase success rates. 
Our approach outperforms others in the one- and few-shot visual imitation learning of pick-and-rearrangement tasks.
In future works, we will investigate local neural descriptors and inter-category generalization of manipulation skills.
%===============================================================================

\def\url#1{}
\bibliographystyle{IEEEtran}
\bibliography{library,HumanoidsGroup}

%===============================================================================

\end{document}